# OSACT4 Shared Task on Offensive Language Detection: Intensive Preprocessing-Based Approach

**Fatemah Husain**
Kuwait University, Department of Information Science, State of Kuwait
f.husain@ku.edu.kw

## Abstract

The preprocessing phase is one of the key phases within the text classification pipeline. This study aims at investigating the impact of the preprocessing phase on text classification, specifically on offensive language and hate speech classification for Arabic text. The Arabic language used in social media is informal and written using Arabic dialects, which makes the text classification task very complex. Preprocessing helps in dimensionality reduction and removing useless content. We apply intensive preprocessing techniques to the dataset before processing it further and feeding it into the classification model. An intensive preprocessing-based approach demonstrates its significant impact on offensive language detection and hate speech detection shared tasks of the fourth workshop on Open-Source Arabic Corpora and Corpora Processing Tools (OSACT). Our team wins the third place (3rd) in the Sub-Task A Offensive Language Detection division and wins the first place (1st) in the Sub-Task B Hate Speech Detection division, with an F1 score of 89% and 95%, respectively, by providing the state-of-the-art performance in terms of F1, accuracy, recall, and precision for Arabic hate speech detection.

## 1. Introduction

Online offensive language detection is one of the most challenging text classification tasks to accomplish due to the ambiguity and informality of the language used in social media platforms. So far, online offensive language detection has been applied to various languages, such as English (Kwok and Wang, 2013; Davidson et al., 2017; Nobata et al., 2016; Pitsilis, Ramampiaro, Langseth, 2018), German (Kent, 2018; Wiedemann et al., 2018), Urdu (Mustafa et al., 2017), Turkish (Özel et al., 2017), Hindi (Bohra et al., 2018; Kapoor et al., 2018), Danish (Derczynski, 2019), and Arabic (Abozinadah, Mbaziira, and Jones, 2015; Mubarak, Darwish, and Magdy, 2017; Alakrot, Murray, and Nikolov, 2018; Mohaouchane, Mourhir, and Nikolov, 2019). Regardless of the text language, a standard text classification pipeline consists of preprocessing, feature extraction, feature selection, and classification model. The preprocessing phase is the one that is the most distinguishable phase among the others based on the text language. Each language contains unique structures and rules, which need to be addressed using unique methods. We develop an intensive preprocessing-based classification model for Arabic offensive language detection. Among the participants of the fourth workshop on Open-Source Arabic Corpora and Corpora Processing Tools (OSACT) throughout the shared tasks, our team wins third place in Sub-Task A Offensive Language Detection division and wins first place in Sub-Task B Hate Speech Detection division.

Text that contains some form of abusive behavior, exhibiting actions with the intention of harming others, is known as offensive language. This abusive behavior could lead to disturbances, disrespect, harm, insults, and anger, thus affecting the harmony of conversations. Wiedemann et al. (2018) describe offensive language as "threats and discrimination against people, swear words or blunt insults" (p.1). Hate speech, aggressive content, cyberbullying, and toxic comments are all different forms of offensive content (Schmidt & Wiegand, 2017). Figure 1 shows an example of an offensive tweet.

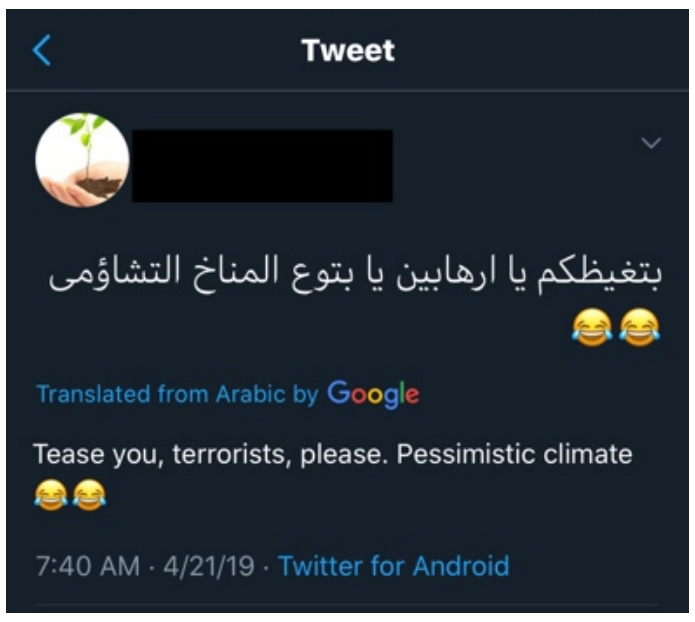

Figure 1: Example for an offensive tweet.

Hate speech is one of the most common forms of offensive language. A text that is targeted towards a group of people-with the intent to cause harm, violence, or social chaos is known as hate speech (Derczynski, 2019). Davidson et al. (2017) define hate speech as "a language that is used to express hatred towards a targeted group or is intended to be derogatory, to humiliate, or to insult the members of the group" (p.1). Figure 2 illustrates an example of a hate speech tweet.

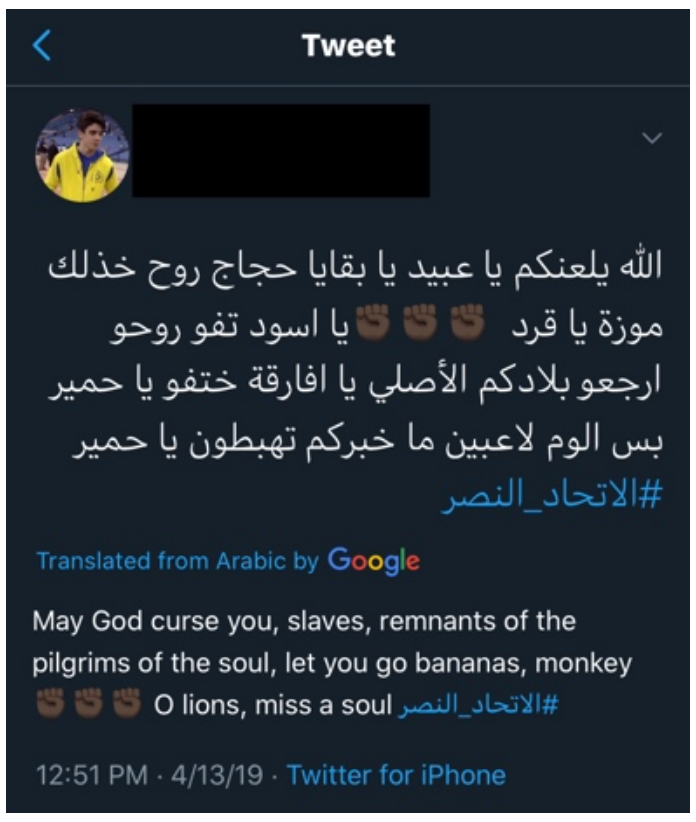

Figure 2: Example for a hate speech tweet.

Developing classification models, which can automatically detect offensive language and hate speech, is a very challenging task due to the following factors: a) the informality of the language used in posts from social media shows posts are usually written using symbols, short forms, and slang that are difficult to semantically process and understand by algorithms; b) the variation and the diversity of the Arabic language dialects and forms; and c) the small sample size of offensive samples; for example, in the dataset used in this study, only 5% of the tweets are labeled as hate speech and 19% of the tweets are labeled as offensive. We apply multiple preprocessing techniques to address the challenges of Arabic offensive language detection. The preprocessing techniques include conversion of emoticons and emoji, conversion of hashtags, normalizing different forms of Arabic letters, normalizing Arabic dialects to Modern Standard Arabic (MSA), normalizing words by categorization, and basic cleaning processes. These intensive preprocessing techniques report valuable influence on the system performance. We train a Support Vector Machine (SVM)-based classifier using character-based count vectorizer (2-5 characters). Results report an F1 score of 89.82% for the offensive language detection model and 95.16% for the hate speech detection model.

In the rest of this paper, we organize the content as follows: Section 2 discusses related work of Arabic offensive language detection on social media, Section 3 introduces data description, details of preprocessing, and the methodology of our models, and experimental results are discussed in Section 4. We also present the conclusion of our work at the end of the paper.

## 2. Related Work

There are multiple studies investigating offensive Arabic tweets to identify abusive Twitter accounts (Abozinadah, Mbaziira, & Jones, 2015; Abozinadah & Jones, 2017; Abozinadah, 2017). Abozinadah, Mbaziira, and Jones (2015) construct an initial dataset starting with 500 Twitter accounts based on a set of Arabic swear words. Then, they check the most recent 50 tweets, profile pictures, and hashtags for each of these 500 Twitter accounts in order to reach a dataset of 350,000 Twitter accounts and 1,300,000 tweets with balanced classes; half were labelled abusive while the other half were labelled as non-abusive. Next, they use three types of features, including profile-based features, tweet-based features, and social graph features to train three classifiers: Naive Bayes (NB), SVM, and Decision Tree (J48). The results show that the NB-based classifier outperforms the other classifiers when used with 100 features and 10 tweets for each account with an accuracy score of 85% (Abozinadah, Mbaziira, & Jones, 2015; Abozinadah, 2017).

Another approach for detecting offensive language was adopted by Mubarak, Darwish, and Magdy (2017) for the purpose of detecting vulgar and pornographic obscene speech in Arabic social media by applying a simple obscene phrases list-based approach. They use a Twitter dataset consisting of 175 million tweets to extract a list of seed words for obscene phrases through manual assessment. Then, they utilize the list to construct 3 sub-lists of obscene words, phrases, and hashtags using multiple measurements, such as the Log Odds Ratio (LOR) for unigrams and bigrams. Intrinsic and extrinsic evaluations were used to evaluate these lists. The intrinsic evaluation consists of manual coding for a list of 100 words that are randomly selected from each list to be marked as either obscene or not. The extrinsic evaluation consists of recall, precision, and F1 measures using a dataset that they developed for the purpose of this evaluation. They then select 10 Egyptian Twitter users from the top controversial users. After that, they extract 100 tweets with at least 10 replies for each user. The final dataset has 100 original tweets and 1,000 replies tweets. Each tweet along with its replies were submitted to CrowdFlower to become coded by three annotators from Egypt using three classes: obscene, offensive, and clean; the inter-annotator agreement is 84%. As a result, they develop a linear match model for each labeled tweet; for example, if a match with a phrase from the list occurs, then, it will then predict a label of obscenity to that tweet. The results among all lists show a highest F1 score of 60%, which demonstrate that a list-based approach is very limited and not a good choice for an obscene detection system.

Arabic language has also been studied also by Alakrot, Murray, and Nikolov (2018a, 2018b) for automatic detection of offensive language. They construct a dataset from YouTube comments based on selecting channels that have controversial videos about celebrities. Their final dataset includes 167,549 comments posted by 84,354 users, and 87,388 replies posted by 24,039 users from 150 YouTube videos (Alakrot, Murray, & Nikolov, 2018a). Two labels were used for the classes: positive to label offensive comments and negative to label ones that are not offensive (Alakrot, Murray, & Nikolov, 2018a). They train an SVM-based classifier using two features: character n-gram (n= 1-5) and word-level features. The results show the best performance when using the SVM-based classifier with 10-fold cross validation and word-level features with 90.05% of an accuracy score (Alakrot, Murray, & Nikolov, 2018b).

Mohaouchane, Mourhir, and Nikolov (2019) explore multiple deep learning models to classify offensive Arabic language for YouTube comments using the same dataset developed by Alakrot, Murray, and Nikolov (2018a). They create 300-dimension word embedding using AraVec, which is an Arabic word embedding tool, trained on Twitter dataset and skip-gram model. Four deep learning models were evaluated for classifying offensive comments, including Convolutional Neural Network (CNN), Bidirectional Long Short-Term Memory (Bi-LSTM), Bi-LSTM with an attention mechanism, and a combined model of CNN and LSTM. The results demonstrate an overall better performance for CNN with a highest accuracy score of 87.84%, a precision score of 86.10%, and an F1 score of 84.05%, while the combined CNN-LSTM model shows a better recall score of 83.46% (Mohaouchane, Mourhir, & Nikolov, 2019).

To our knowledge, the only Arabic hate speech detection studies are the studies of Albadi, Kurdi, and Mishra (2018, 2019) and the study of Chowdhury et al. (2019), which investigate religious hate speech in Arabic language for Twitter data, both using the same dataset. Albadi, Kurdi, and Mishra (2018) develop a logistic regression-based model and an SVM-based model using a character n-gram feature (n= 1 to 4), and a Gated Recurrent Unit (GRU) based on the Recurrent Neural Network (RNN) with the Twitter Continuous Bag-of-Word (CBOW) 300-dimension embedding model provided by AraVec, batches of size 32, and Adam as the optimizer. Their findings have indicated that for some religious minorities in the Middle East — Jews, Atheists, and Shia— almost half of the tweets that were mentioning these minorities were referring to them within a hate speech content (Albadi, Kurdi, & Mishra, 2018). Furthermore, results report best performance when using the GRU-based model with an F1 score of 77% (Albadi, Kurdi, & Mishra, 2018). Albadi, Kurdi, and Mishra (2019) enhance the same GRU-based model with additional temporal, users, and content features in another study, then report the state-of-the-art performance in terms of a recall score of 84%. Chowdhury et al (2019) extend previous studies done by Albadi, Kurdi, and Mishra (2018, 2019) to investigate the effects of community interactions and social representations in detecting religious hate speech in the Arabic Twitter sphere. They use multiple features, including word embedding, node embedding, sentence representation, and character n-gram features (n = 1 - 4). Several classification models were explored using multiple combinations of features, such as GRU, logistic regression, SVM, LSTM, Bi-LSTM, CNN and Bi-GRU in addition to combining multiple models, using self-attention mechanisms, and using Node2Vec criteria. Therefore, results have shown a high accuracy score of 81% that was obtained by the model containing a combination of Bi-GRU, CNN, and NODE2VEC. While displaying the best F1 score, recall and precision were recorded by the model that combined LSTM, CNN and NODE2VEC: 89%, 78%, and 86% respectively.

Previous studies focus on features extraction and classification model, while we focus more on text preprocessing. The preprocessing steps we follow in this study are not identical to any of the preprocessing steps of the previous studies.

## 3. Dataset and Methodology

### 3.1 Dataset Description

The shared task of the fourth workshop on Open-Source Arabic Corpora and Corpora Processing Tools (OSACT) in Language Resources and Evaluation Conference (LREC) 2020 provides Twitter dataset for offensive language detection in Arabic language. The main goal of this shared task is to identify and categorize Arabic offensive language in Twitter. The organizers collect tweets through Twitter API and annotate them hierarchically regarding offensive language and offense type. The task is divided into two sub-tasks: a) detecting if a post is offensive or not offensive; and b) identifying offensive content type of an offensive post as hate speech or not hate speech.

The provider of the dataset performs some preprocessing to ensure the privacy of users. Twitter user mentions were substituted by “@USER”, URLs had simply been substituted by “URL”, and empty lines were replaced by “<LF>”.

The shared task issues the dataset in three different parts, training dataset, development dataset and testing dataset. The summary of datasets distribution is presented in Table 1. Training dataset and development dataset are provided with their actual labeled, while the testing dataset consists of 2,000 unlabeled tweets for competition evaluation purposes. The training dataset consists of 6,839 tweets with 1,371 offensive tweets and 350 hate speech tweets. The development dataset consists of 1,000 tweets with 179 offensive tweets and 44 hate speech tweets.

| Labels | Training | Development | Testing |
|---|---|---|---|
| Not Offensive | 5,590 | 821 | * |
| Offensive | 1,410 | 179 | * |
| Not Hate Speech | 6,639 | 956 | * |
| Hate Speech | 361 | 44 | * |
| Total Tweets | 7,000 | 1,000 | 2,000 |

Table 1: Datasets distribution (* not available)

We explore multiple characteristics of the dataset including most frequent emoji, emoticons, and words. The most frequent emoticons and emoji are very similar in both classes. Table 2 and Table 3 show the top 5 most frequent emoji and emoticons for both not offensive and offensive classes, respectively. Most frequent words are also very similar in both classes. Figure 3 and Figure 4 show the most frequent words for both not offensive class and offensive class, respectively.

| Emoji | % | Emoticon | % |
|---|---|---|---|
| 😂 | 18% | :< | 44% |
| ❤ | 12% | :) | 23% |
| 💙 | 5% | :( | 16% |
| 💛 | 5% | ;D | 6% |
| ♥ | 4% | :D | 6% |

Table 2: The top 5 most frequent emoji and emoticons for not offensive tweets

| Emoji | % | Emoticon | % |
|---|---|---|---|
| 😂 | 38.87% | ;D | 40% |
| 🤣 | 8.52% | :< | 33.33% |
| ❤ | 3.83% | :( | 13.33% |
| 😭 | 3.74% | :) | 6.67% |
| 💙 | 2.43% | d: | 6.67% |

Table 3: The top 5 most frequent emoji and emoticons for offensive tweets

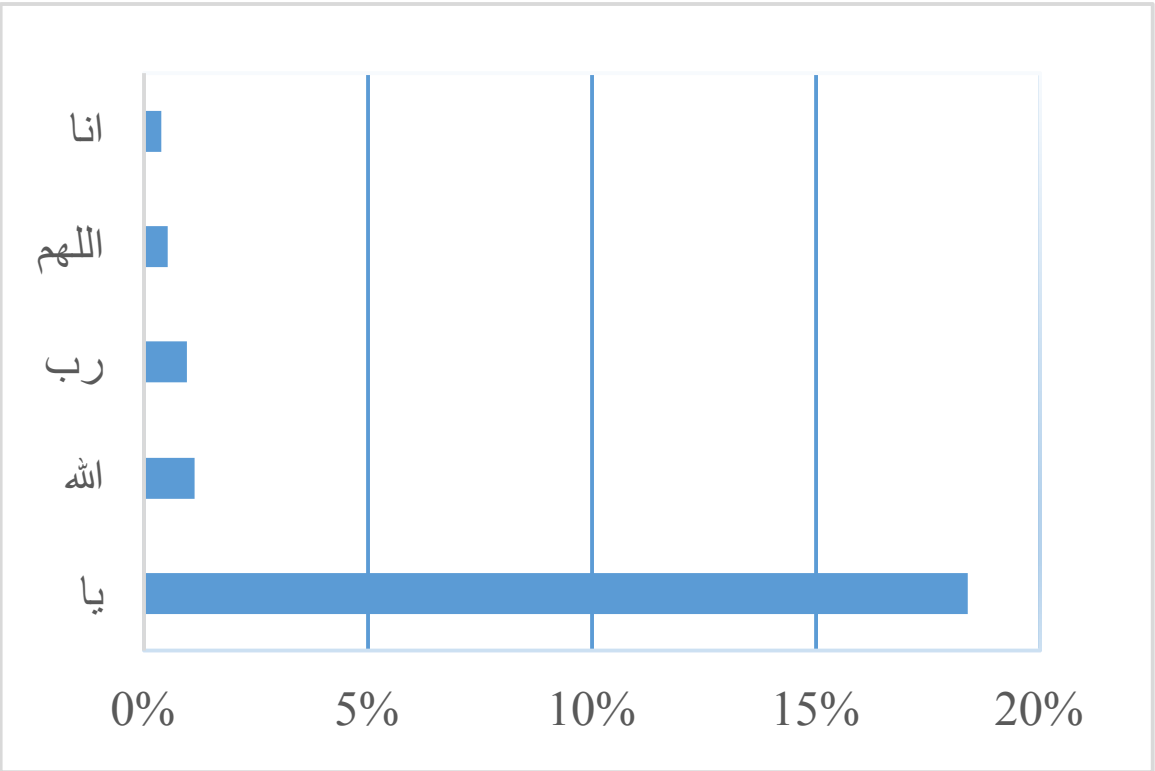

Figure 3: Most frequent 5 words in not offensive tweets

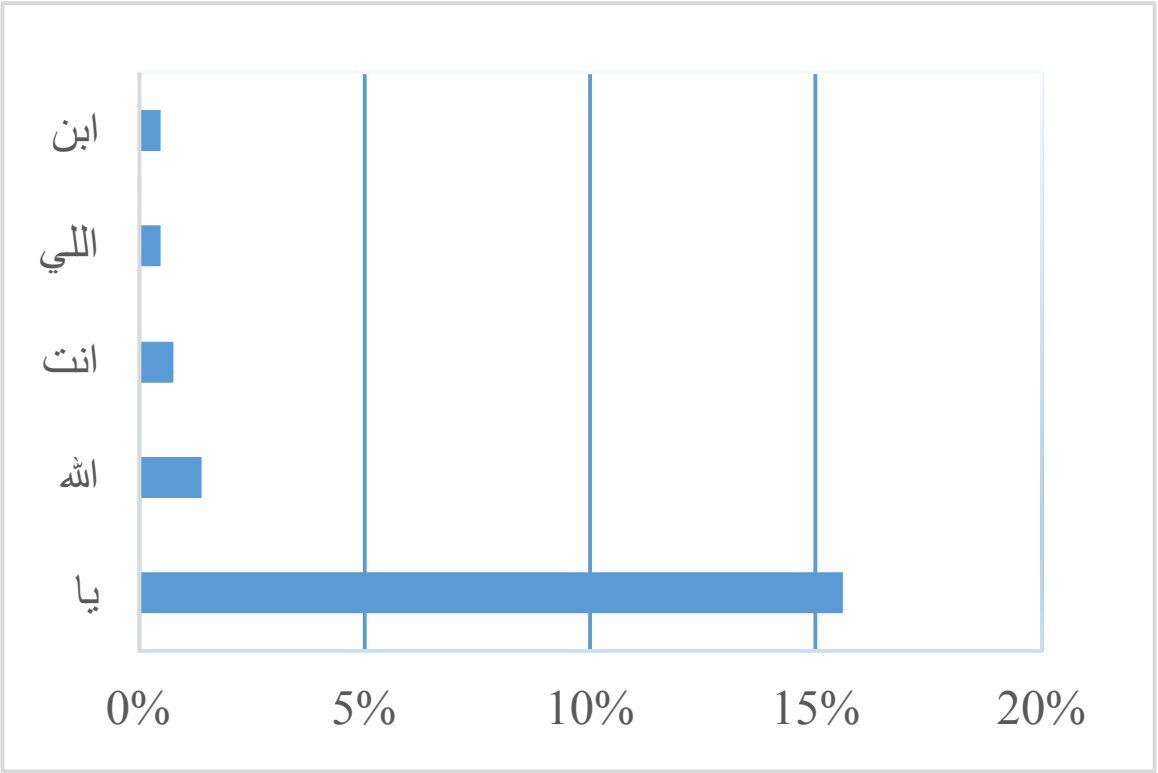

Figure 4: Most frequent 5 words in offensive tweets

## 3.2 Preprocessing

### 3.2.1 Emoji and Emoticon Conversion

Emoji and emoticons are often used to convey feelings and attitude, which is very valuable to the task of offensive detection. Knowing some contextual information about the author of the post can provide insight into the textual content of the post. Moreover, a previous study converts an emoji to a textual label, and uses it to provide sentimental features to train a classifier for aggression detection, thus reporting better performance than the other model that does not consider emoji conversion (Orasan, 2018). Thus, we consider converting emoji and emoticons to an Arabic textual label that describe the content of them. To have a more robust system that can scale and cover more emoji and emoticons, we then extract the entire set of emoji defined by Unicode.org (Unicode Organization, n.d.). Beautifulsoup4 4.8.2, a Python package for parsing HTML and XML documents, was used to scrape the emoji list available at the Unicode Organization website. We extract the Unicode and name of each emoji using Beutifulsoup4. The emoji list contains 1,374 emoji. The Unicode Organization website provides textual descriptions for each emoji written in English. We extract these textual descriptions, then, we use Translate 1.0.7 python package to translate them into Arabic language. Thus, during preprocessing of the data, each emoji is converted to its Arabic description. For example, 😜 is replaced by "وجه يغمز مع لسان" . In addition, we manually construct an emoticons list that includes a total of 140 emoticons with their textual descriptions in Arabic language. For example, ":-X" is replaced by " معقود اللسان ". Next, we analyze the description phrase as a regular textual phrase in tweets so that it could maintain its semantic meaning after removing the original emoji and emoticon.

### 3.2.2 Arabic Dialects Normalization

The Arabic dialects have various forms, based on geography and social class (Habash, 2010). Arabic dialects are the Arabic languages that have often been used in user-generated contents such as Twitter. Habash (2010) categorizes the Arabic dialects into seven dialects: Gulf, Egyptian, Iraqi, North African, Yemenite, Levantine, and Maltese Arabic, which is not always considered one of the Arabic dialects. It is very crucial to consider the variations among different dialects when detecting offensive language, as some words that have exactly the same pronunciation and spelling might have different meanings. For example, the word "عافية - Afiah" means health in Gulf, Egyptian, Iraqi, and Levantine dialects ; however, it also means fire in Moroccan Arabic. We try to solve the variation in dialects by a dimensionality reduction approach. We reduce the dimensionality of the data by normalizing the variation of dialects for a set of nouns to be converted from dialectal Arabic to MSA. For example, the variations of the word boy, "رجل", "زلمة", and "زول" are converted to "ولد". The set of nouns is manually constructed based on manual insepction for a sample of the dataset and based on our own experience as native Arabic speakers.

### 3.2.3 Words Categorization

Normalizing and reducing the dimensionality can improve the performance of the model. From our manual inspection for a sample of tweets from the dataset, we notice that it is very common to mention name of animals among the hate speech tweets. Thus, we manually create a list of the most common animal names used in different Arabic dialects, such as "كلب - dog", "خنزير - pig", "حية - snake". The list considers the variation in dialects in animal names, such as the word cat in MSA is "Qetta/قطة ", while in Egyptian it is "Otta/ أطة", in Levantine it is "Bisse/بسة", in Gulf it is "Qatwa/قطوة", in Moroccan it is "Qetta/قطة ", in Iraqi it is "Bazzuna/بزونة", and in Yamani it is "demah/دمة". In addition to the variation in dialects, the list includes variations of the same animal names, such as the female name, male name, plural name. Thus, for the word cat, we include singular female name "Qetta/قطة ", singular male name "Qett/قط ", two female name "Qettatan/قطتان ", two male name "Qettan/قطان ", and plural name "Qettat/قطط ". The list contains a total of 335 lexicons. All animal names that are listed and occurred in the dataset were converted to the word animal in Arabic "حيوان". Accordingly, all animal names that are included in tweets are reduced to only one word.

### 3.2.4 Letters Normalization

Arabic letters can be written in various format depending on the location of the letter within the word. We normalize Alif (أ،آ،إ to ا ), Alif Maqsura (ئ،ى to ي), and Ta Marbouta (ة to ه). Letters that were repeated more than two times within a word were reduced to two times only.

### 3.2.5 Hashtag Segmentation

Hashtags are commonly used in Twitter to highlight important phrases within the tweet. Thus, it is very important to consider hashtags during the preprocessing phase to convert hashtags into a meaningful format. We remove the "#" symbol and replace "_" by a space. For example, the hashtag "#الهلال_التعاون" is converted to "الهلال التعاون", which is easier for the system to understand and process.

### 3.2.6 Miscellaneous

Tweets were filtered to remove numbers, kashida, HTML tags, more than one space, three or more repetitions of any character, and some symbols or terms (e.g., " _", "", "\"", "'s", "...", "!", "?", "I", "@USER", "USER", "URL", ".", ";", ":", "/", "\\", ", ", "#", "@", "$", "&", ")", "(", "\"). We borrow the list of Arabic stopwords defined by Alrefaie

(2016), which contains 750 stopwords. Furthermore, we remove diacritics that were used in tweets containing text from the holy Qur'an or poetry.

### 3.2.7 Upsampling

As can be noticed from Table 1, the classes are highly imbalanced, with only 5% hate speech tweets and 19% offensive tweets. we try to solve the problem with the imbalanced distribution of classes by using up-sampling. The original training dataset has 7,000 tweets with a first label hierarchy of 1,410 offensive tweets and 5,590 not offensive tweets. After up-sampling, the number of tweets for each class, offensive tweets and not offensive tweets, becomes 5,590. The second label hierarchy is sharply imbalanced, the original training dataset consists of 361 hate speech tweets and 6639 not hate speech tweet. After up-sampling, the number of tweets for each class, hate speech tweets and not hate speech tweets, becomes 6,639. We use resample function from Python scikit-learn library to implement upsampling. Table 4 illustrates example of tweets before preprocessing and after preprocessing.

| Tweet Before Cleaning | Tweet After Cleaning |
|---|---|
| يا عمرررررررري يا بو شعر ثاير 😭😭😭😭 💛💛💛💛 | ىا عمررى ىا بو شعر ثاىر وجه البكاء بصوت عال وجه البكاء بصوت عال وجه البكاء بصوت عال وجه البكاء بصوت عال قلب اصفر قلب اصفر قلب اصفر قلب اصفر |
| @USER: بطلو انانيه يا شعب يا انصرافي ويلا على صفوف عيشكم وبنزينكم | بطلو انانىه ىا شعب ىا انصرافى وىلا صفوف عىشكم وبنزىنكم |
| RT @USER: ما كان ممكن تبدأ بيه بدل ما تخسر تبديل يا حمار يابن الحمار يا غبيييي | ممكن تبدا بىه بدل تخسر تبدىل ىا حىوان ىابن حىوان ىا غبىى |

Table 4: Examples of tweets before preprocessing and after preprocessing

## 3.3 Feature Extraction

We use two features in training the models, which consists of character-based count vectorizer and TF-IDF vectorizer, both with 2 to 5 characters. Previous studies highlight the importance of using character-based features over word-based features for offensive language detection because it is a language independent feature that can work with misspelling errors or obfuscating offensive words, which are commonly practiced on social media posts (Bohra et al., 2018; Nobata et al., 2016). Both features are implemented using Python scikit-learn library.

## 3.4 Methodology

### 3.4.1 Preliminary Models

We explore the effect of each preprocessing technique on the performance of the model before applying them to the final models. Previous studies on offensive language detection report high performance when using an SVM-based classifier (Abozinadah and Jones, 2017; Schmidt and Wiegand, 2017; Albadi, Kurdi, and Mishra, 2018). The SVM classifier focuses on maximizing the margin, the distance of the closest points to the hyperplane that separate between instances of classes using a linear function (Goodfellow, Bengio, and Courville, 2016). Thus, we decide to be in-line with the findings from earlier studies and use the linear SVM-based classifier in this step trained using the first label hierarchy of the dataset; offensive or not offensive, for the purpose of this exploration task. We did not use the testing dataset during this step. The training dataset used to train the models and the development dataset used to evaluate the models. During this preliminary modeling phase, two main goals were targeted. The first goal is to identify if count vectorizer outperforms TF-IDF vectorizer or not. To accomplish this goal, we trained two SVM-based models using the raw dataset without performing any preprocessing technique, one model applies the count vectorizer and the other one applies the TF-IDF vectorizer. Results are shown in Table 5. The TF-IDF vectorizer is 2% better than the count vectorizer in precision score. However, having a comprehensive measurement that consider multiple factors in evaluating the model is very important. The dataset is highly imbalanced, so we consider F1 score in evaluating the performance rather than accuracy. The accuracy score is often gives misleading results in similar situation with imbalanced dataset. F1 score is the harmonic mean for recall and precision. The count vectorizer outperforms the TF-IDF vectorizer by 2% in F1 score. This finding is consistent with Wiedeman et al. (2018) finding, which reports the unsuitability of TF-IDF vectorizer over twitter datasets because tweets are very short, making it not an optimal source for IDF estimations. Thus, we consider adopting the count vectorizer as the feature for the rest of the models.

| Feature | Precision | Recall | F1 | Accuracy |
|---|---|---|---|---|
| Count Vectorizer | 85% | **76%** | **79%** | **89%** |
| TF-IDF Vectorizer | **87%** | 74% | 77% | 88% |

Table 5: Performance results for preliminary models based on features

The second goal of this preliminary modeling phase is to evaluate each preprocessing technique separately and measure their effects on the performance. Table 6 presents the results from the preliminary preprocessing exploration models. Only two preprocessing techniques were merged; hashtags segmentation and miscellaneous cleaning processes due to their relatedness and similarity in term of platform specific attributes. Preliminary results illustrate the contribution of each preprocessing technique to the performance of the model; however, we also expect to have different results when used on larger dataset and when used for hate speech detection. The results report the highest contribution in term of F1 score to letters normalization, followed by dialect normalization and word categorization, then, emoji and emoticon conversion, miscellaneous cleaning and hashtag segmentation, and finally, upsampling. The results demonstrate some issues with the upsampling technique, it reduces F1 score from 79%, as it shown in Table 5, to 71%. Consequently, we decide to apply all the preprocessing techniques except upsampling for the main models.

| Preprocessing | Precision | Recall | F1 | Accuracy |
|---|---|---|---|---|
| Emoji and Emoticon | 83% | 78% | 80% | 89% |
| Dialects Normalization | **85%** | 80% | 82% | **90%** |
| Word Categorization | 84% | 80% | 82% | **90%** |
| Letters Normalization | **85%** | **81%** | **83%** | **90%** |
| Miscellaneous and Hashtags Segmentation | 82% | 76% | 79% | 89% |
| Upsampling | 69% | 75% | 71% | 80% |

Table 6: Preliminary performance evaluation results from each preprocessing technique on offensive detection

### 3.4.2 Baseline Models

The baseline models include two linear SVM-based models, both trained on dataset without any sort of preprocessing technique and using a count vectorizer with 2 to 5 characters. The first baseline model classifies tweets to either Offensive (OFF) or Not Offensive (NOT_OFF), while the second one classifies tweets to either Hate Speech (HS) or Not Hate Speech (NOT_HS). To assess our goal in investigating the effect of preprocessing on offensive detection and hate speech detection, we use the same feature and classifier for baseline models and main models. All Models are implemented using Python scikit-learn library.

### 3.4.3 Sub-Task A Model : Offensive Language Detection

The main model for offensive language detection, which classifies tweets to either offensive (OFF) or not offensive (NOT_OFF), is a Linear SVM-based classifier. The model is trained on full preprocessed tweets that have been preprocessed using all techniques mentioned earlier; conversion of emoticons and emoji, conversion of hashtags, normalizing different forms of Arabic letters, normalizing Arabic dialects to MSA, normalizing words by categorization, and other miscellaneous cleaning processes. As we mentioned earlier, the feature used in training the model is the character-based count vectorizer with 2 to 5 characters. The model is implemented using Python scikit-learn library.

### 3.4.4 Sub-Task B Model : Hate Speech Detection

The same exact model of Sub-Task A is used for Sub-Task B, including the same preprocessing and feature extraction techniques. However, the model classifies tweets to either hate speech (HS) or not hate speech (NOT_HS).

## 4. Experiment Results

Table 7 shows the results of performance evaluation for the baseline model and the main model for offensive language detection task, and Table 8 shows the results for hate speech detection task. The baseline models were evaluated using the development dataset while the main models were evaluated using the testing dataset through the shared task competition evaluators. The dataset is highly imbalanced, thus, the accuracy score might not be very informative to evaluate the performance. The F1 score increased from 79% for the baseline model before preprocessing to 89% after preprocessing for Sub-Task A for offensive detection, which was ranked the 3rd place. For Sub-Task B for hate speech detection, F1 score increased sharply from 67% to 95%, which was ranked the 1st place.

The most noticeable point from the tables is that in Sub-Task B results are better than in Sub-Task A, given the fact that the class distribution is more skewed than that of Sub-Task A and the number of training instance is much smaller than Sub-Task A. This interesting point demonstrates how data preprocessing adds value to the performance, even for hate speech class that is very rare.

| Model | Precision | Recall | F1 | Accuracy |
|---|---|---|---|---|
| Baseline Model | 85% | 76% | 79% | 89% |
| Main Model | 89% | 90% | 89% | 90% |

Table 7: Performance evaluation of offensive language detection task

| Model | Precision | Recall | F1 | Accuracy |
|---|---|---|---|---|
| Baseline Model | 74% | 63% | 67% | 96% |
| Main Model | 95% | 95% | 95% | 95% |

Table 8: Performance evaluation of hate speech detection task

## 5. Conclusion

The preprocessing phase is one of the key phases within the text classification pipeline, including offensive language and hate speech classification. The ambiguity and informality of social media language increase the complexity of achieving high classification performance, particularly for Arabic text that has multiple dialects. Our study shows the competitive results obtained for offensive language detection and hate speech detection by using intensive preprocessing techniques to filter and clean the dataset before feeding it into the rest of the phases for the text classification pipeline. Moreover, results report the significant impacts of preprocessing on a very challenging task, such as the hate speech classification, with very small sample size of 5% from the overall dataset.

In the future, we hope to enhance the available studies of offensive language detection and hate speech detection by investigating our preprocessing techniques using some deep learning models. Another future direction is to use more advanced features in training the model, such as word embedding.

## 6. Bibliographical References